\documentclass[conference]{IEEEtran}
\usepackage{tabularx}
\usepackage{array, boldline, booktabs, makecell}
\usepackage{multirow}
\usepackage{amsmath,amssymb,amsfonts}
\usepackage{algorithmic}
\usepackage{graphicx}
\usepackage{textcomp}
\usepackage{caption}
\usepackage{xcolor}
\usepackage{hyperref, etoolbox}
\usepackage{float}
\usepackage{bm}
\def\BibTeX{{\rm B\kern-.05em{\sc i\kern-.025em b}\kern-.08em
    T\kern-.1667em\lower.7ex\hbox{E}\kern-.125emX}}
\usepackage{algorithm}
\usepackage{algorithmic}

\usepackage{newfloat}
\usepackage{listings}

\usepackage{amsmath}
\usepackage{amsthm}
\usepackage{booktabs}
\usepackage{algorithm}
\usepackage{algorithmic}
\usepackage{multirow}

\usepackage{mathrsfs}

\newcommand*{\Scale}[2][4]{\scalebox{#1}{$#2$}} 

\usepackage{booktabs}
\usepackage{tabularx, ragged2e}
\usepackage{verbatim}
\usepackage{fancyvrb}
\usepackage{fvextra}

\usepackage{amsmath,amsfonts,bm}

\def\eqref#1{(\ref{#1})}
\def\Eqref#1{Equation~(\ref{#1})}

\def\1{\bm{1}}

\def\vp{{\bm{p}}}

\def\vx{{\bm{x}}}

\DeclareMathAlphabet{\mathsfit}{\encodingdefault}{\sfdefault}{m}{sl}
\SetMathAlphabet{\mathsfit}{bold}{\encodingdefault}{\sfdefault}{bx}{n}

\def\gD{{\mathcal{D}}}

\def\gL{{\mathcal{L}}}

\def\gX{{\mathcal{X}}}
\def\gY{{\mathcal{Y}}}

\def\sE{{\mathbb{E}}}

\newcommand{\E}{\mathbb{E}}
\newcommand{\Ls}{\mathcal{L}}

\DeclareMathOperator*{\argmin}{arg\,min}

\begin{document}

\title{End-to-End Anti-Backdoor Learning on Images and Time Series}

\author{\IEEEauthorblockN{Yujing Jiang\textsuperscript{1}, Xingjun Ma\textsuperscript{2}, Sarah Monazam Erfani\textsuperscript{1}, Yige Li\textsuperscript{3}, James Bailey\textsuperscript{1}}
\IEEEauthorblockA{\textsuperscript{1}\textit{Faculty of Engineering and Information Technology, The University of Melbourne}\\
\{yujingj@student., sarah.erfani@, baileyj@\}unimelb.edu.au}

\IEEEauthorblockA{\textsuperscript{2}\textit{School of Computer Science, Fudan University}\\
xingjunma@fudan.edu.cn}

\IEEEauthorblockA{\textsuperscript{3}\textit{School of Computer Science and Technology, Xidian University}\\
yglee@stu.xidian.edu.cn}
}

\maketitle

\begin{abstract}

Backdoor attacks present a substantial security concern for deep learning models, especially those utilized in applications critical to safety and security. These attacks manipulate model behavior by embedding a hidden trigger during the training phase, allowing unauthorized control over the model's output during inference time. Although numerous defenses exist for image classification models, there is a conspicuous absence of defenses tailored for time series data, as well as an end-to-end solution capable of training clean models on poisoned data.
To address this gap, this paper builds upon Anti-Backdoor Learning (ABL) and introduces an innovative method, End-to-End Anti-Backdoor Learning (E2ABL), for robust training against backdoor attacks. Unlike the original ABL, which employs a two-stage training procedure, E2ABL accomplishes end-to-end training through an additional classification head linked to the shallow layers of a Deep Neural Network (DNN). This secondary head actively identifies potential backdoor triggers, allowing the model to dynamically cleanse these samples and their corresponding labels during training. Our experiments reveal that E2ABL significantly improves on existing defenses and is effective against a broad range of backdoor attacks in both image and time series domains. 
\end{abstract}

\section{Introduction}


Deep learning has achieved remarkable performance in computer vision tasks such as object detection \cite{redmon2016you}, motion tracking \cite{balasundaram2017vision}, and autonomous driving \cite{cordts2016cityscapes}, as well as time series analysis in fields like finance \cite{peia2015finance}, smart manufacturing \cite{essien2020deep}, and healthcare \cite{penfold2013use}. With the increasing deployment of Deep Neural Networks (DNNs) in real-world applications, their vulnerability to backdoor attacks has become a significant concern. Backdoor attacks occur either by poisoning a few training samples with a trigger pattern \cite{gu2017badnets} or by manipulating the training procedure \cite{liu2020reflection}, thereby implanting a backdoor in the DNN model. The compromised model learns a strong correlation between the trigger pattern and a chosen backdoor label. At inference time, it predicts correct labels on clean inputs while exhibiting a systematic bias towards the backdoor label in the presence of the trigger. This issue is especially serious as these technologies are being deployed in safety-critical applications. Therefore, developing defenses against such backdoor attacks is becoming a critical necessity.

Backdoor attacks generally have two primary objectives: 1) high effectiveness, i.e., high attack success rate (ASR), and 2) high stealthiness, i.e., maintaining a high clean accuracy (CA) while visually undetectable. High effectiveness allows the attacker to manipulate the model's prediction in a more precise manner, while high stealthiness ensures that the attacks cannot be easily detected by rudimentary filtering or manual inspection. Stealthiness also involves designing subtle trigger patterns that do not impact the model's clean performance (performance on clean data), making detection even more challenging. Several works \cite{gu2017badnets,liu2020reflection,turner2019label,cheng2021deep,ding2022towards,jiang2022backdoor} have studied backdoor attacks on both image and time series data. 

As more and more DNNs are being trained and employed in different types of real-world applications, defending against malicious and stealthy backdoor attacks on different tasks and data modalities has become an imperative task. This work takes the first attempt to design one single defense method that could work for two data modalities, i.e., images and time series. The reason why we chose images and time series is that both modalities are continuous (unlike discrete texts), their classification tasks are well-studied, and there exist multiple backdoor attacks for both types of data.
However, current defense methods are mostly tailored for image data and have not been well studied for time series data. It is thus unclear whether defenses developed against image backdoor attacks are suitable for time series backdoor attacks. 


Anti-Backdoor Learning (ABL) \cite{li2021anti} is a robust training method that was initially introduced to train clean models on poisoned data. It has demonstrated promising results against a diverse set of backdoor attacks on image datasets. 
Nonetheless, ABL has some limitations. One notable limitation is its two-stage training process. In the first stage, the model undergoes a training phase for a specified number of epochs, following which a subset of suspected backdoor samples is isolated. The model then enters a secondary training phase aimed at ``unlearning" these potentially harmful patterns. Each of these stages demands distinct training objectives, thereby complicating the process and potentially reducing its efficiency.

In this paper, we propose an End-to-End Anti-Backdoor Learning (E2ABL) training method that is capable of training a clean model on a poisoned dataset in an end-to-end manner. Our approach eliminates the need to change the training objectives or extend training epochs. Specifically, we introduce a secondary classification head attached to the shallow layers of the DNN model. This second head traps potential backdoor samples and corrects their labels. The second head is specifically designed to be sensitive to backdoor correlations and samples. This ensures backdoor samples are captured in the shallow layers, safeguarding the primary head and steering the model training toward a more secure and trustworthy direction.

To summarize, our main contributions are:

\begin{itemize}
\item We introduce End-to-End Anti-Backdoor Learning (E2ABL), a novel end-to-end robust training method that can train backdoor-free models on backdoored datasets. E2ABL works effectively for both image and time series data, and to the best of our knowledge, is the first backdoor defense method for time series models.

\item E2ABL proposes a novel strategy of using a second head to safeguard the learning of the main network against backdoor attacks. The second head is designed to learn, capture, and rectify backdoored samples in real-time, and is trained concurrently with the main head to neutralize the impact of backdoor attacks.

\item Through extensive empirical evaluations, we demonstrate that E2ABL can serve as an effective backdoor defense method against a broad spectrum of backdoor attacks on both image and time series data. Further, models trained using our E2ABL consistently outperform those trained by other defense methods, exhibiting higher clean accuracy (CA) and lower attack success rate (ASR).
\end{itemize}

\section{Related Work}

In this section, we provide a brief overview of the existing literature focusing on both backdoor attacks and defenses.

\subsection{Backdoor Attack}

\subsubsection{Image Attacks}
Backdoor attacks optimize two primary objectives including attack effectiveness, and stealthiness. These objectives are fulfilled by optimizing metrics such as the attack success rate and clean accuracy, while also focusing on the design of increasingly subtle and inconspicuous trigger patterns. Additionally, efforts are made to minimize the rate at which training samples are poisoned. The seminal work in this domain, BadNets \cite{gu2017badnets}, laid the foundation for backdoor attacks on Deep Neural Networks (DNNs) by introducing a simplistic checkerboard pattern affixed to the lower-right corner of a clean image. In the wake of this pioneering study, subsequent research has ventured into more sophisticated techniques, such as the integration of blended backgrounds \cite{chen2017targeted}, the inclusion of natural reflections \cite{liu2020reflection}, and the utilization of imperceptible noise \cite{liu2020reflection,turner2019label,cheng2021deep,bagdasaryan2021blind}. There are also works utilizing adversarial patterns \cite{zhao2020clean} and sample-wise patterns \cite{nguyen2020input,li2021invisible} as backdoor attack methods.
Remarkably, some attacks have even demonstrated the ability to reverse-engineer training data without requiring access to the original dataset \cite{liu2018trojaning}. Moreover, clean-label attacks, which insert triggers without altering the actual class labels, have gained attention in recent research \cite{zhao2020clean,shafahi2018poison,turner2019clean,zhu2019transferable,saha2020hidden}. Many of these methods achieve considerable attack success rates while contaminating less than 10\% of the training dataset, some being effective at a surprisingly low poisoning rate of 0.1\%. These trends underscore the stealthy and evasive nature of backdoor attacks, thereby accentuating the urgent need for robust anti-backdoor learning mechanisms. It is worth noting that the majority of these attacks and subsequent studies have been primarily focused on image data and image classification models.

\subsubsection{Time Series Attacks}
The field of backdoor attacks on time series data is still in its nascent stage. One of the pioneering works in this domain is by \cite{wang2020backdoor}, which transformed time series data into 2D images. This transformation allowed them to apply existing backdoor attack techniques originally designed for image data. Building on this initial exploration, \cite{ding2022towards} introduced TimeTrojan, a specialized backdoor attack tailored for DNN-based time series classifiers. TimeTrojan utilizes a multi-objective optimization framework, enabling it to establish strong and stealthy links between the hidden trigger and the target label, thereby making the attack more effective and less detectable.
More recently, the research landscape has seen the advent of a Generative Adversarial Network (GAN)-based approach presented by \cite{jiang2022backdoor}. This method overlays a unique, sample-specific trigger on each compromised time series data point. The GAN-based approach not only elevates the level of stealthiness but also enhances the natural appearance of the poisoned data, further complicating detection efforts.

In addition to data poisoning-based backdoor attacks, it is worth mentioning another category of attacks that directly target the model's parameters \cite{dumford2020backdooring,garg2020can}. These parameter-based attacks can be executed independently or in conjunction with data poisoning-based strategies, thereby presenting a multi-faceted threat landscape. However, the primary focus of this paper remains on countermeasures against data poisoning-based backdoor attacks. The exploration of defenses against model parameter manipulation-based backdoor attacks constitutes an avenue for our future work, given its distinct set of challenges and implications.

\subsection{Backdoor Defenses (On Image Data)}

While a large number of defense methods have been proposed to combat backdoor attacks on image data, there is a noticeable absence of techniques specifically tailored for time series data. Prominent existing defenses, such as Mode Connectivity Repair (MCR) \cite{zhao2020bridging}, Neural Attention Distillation (NAD) \cite{li2021neural}, Adversarial Neuron Pruning (ANP) \cite{wu2021adversarial}, and Reconstructive Neural Pruning (RNP) \cite{li2023reconstructive} are principally engineered to counter the harmful influence of backdoor triggers in image-based neural networks. These methods largely neglect the unique characteristics and vulnerabilities inherent to time series models. Moreover, earlier defense strategies like Fine-Pruning \cite{liu2018fine} have been found to be less effective in the face of evolving, more sophisticated backdoor attacks \cite{liu2020reflection,yao2019latent}.

In response to these challenges, \cite{li2021anti} introduced the concept of Anti-Backdoor Learning (ABL) where the goal is to train clean models directly on poisoned data. Their proposed ABL method consists of two distinctive stages. In the first stage, the target model undergoes initial training for several epochs, following which a limited number of samples with the lowest loss values are isolated as backdoor samples. The second stage involves fine-tuning the model in conjunction with maximizing the model's loss on the isolated backdoor samples. By employing different training objectives in the two stages, ABL diverges from standard training procedures which are mostly end-to-end training with one single loss.

Despite these advancements in the image domain, the time series domain is notably under-researched. Particularly, there are no established defense methods specifically designed to counter backdoor attacks on time series models, highlighting an urgent gap in the current literature. In this paper, we advocate the concept of ABL and propose a novel End-to-End Anti-Backdoor Learning (E2ABL) method that works for both images and time series. The second head attached to the main network in E2ABL serves as an innovative mechanism that is capable of real-time identification, capture, and rectification of backdoor samples, thereby protecting the integrity of the learning process conducted by the main network.

\section{End-to-End Anti-Backdoor Learning}

This section presents our innovative E2ABL method, which features two key advancements: a dual-head model architecture and a true class recovery mechanism. We also outline the threat model, formulation, and motivation for E2ABL.

\subsection{Threat Model}

In this study, we focus on classification tasks involving both image and time series data. The methodology could potentially be extended to other types of data and applications, such as natural language processing or anomaly detection. We adopt a classic data poisoning-based threat model where the adversary can poison the training data by injecting backdoor trigger patterns into a few clean samples. The backdoor-poisoned dataset is then used by the defender to train a target DNN model. The defender has full control over the training process but has no prior knowledge of the poisoning statistics, including the existence of an attack, the number of poisoned samples, the trigger pattern(s), or the backdoor target label.
The defender's objective is to train a clean, backdoor-free model from a potentially poisoned dataset, aiming to attain a clean performance equivalent to models trained on clean data. This scenario embodies a robust training environment where backdoor mitigation or elimination strategies can still be applied effectively, even for those devised in different settings.

\subsection{Problem Formulation}
\label{sec:prob}

Consider a benign training set comprised of $N$ independently and identically distributed samples, denoted as $\gD=\{(\vx_{i}, y_{i})\}_{i=1}^{N}$. In this dataset, each $\vx_{i}$ represents a single training sample, and $y_{i}$ is its corresponding ground truth label. A classification model represented as $f_{\theta}$, learns the function $f_{\theta}:\gX \rightarrow \gY$, which maps the input space to the label space. This learning process is usually achieved by minimizing the empirical error, as shown below:
\begin{equation}
\label{eq:std_training}
\Ls = \E_{(\vx,y) \sim \gD} [\ell(f_{\theta}(\vx), y)],
\end{equation}
\noindent where $\ell(\cdot)$ denotes the loss function, such as the widely used cross-entropy loss. A backdoor adversary manipulates a portion of the benign dataset $\gD$, creating a backdoor-poisoned subset $\gD_{p}$, while leaving the remaining dataset $\gD_c$ intact. This forms a compromised dataset denoted as $\gD'=\gD_p \cup \gD_c$. The poisoned samples are created through the operation $\vx_{i}'=\vx_{i} \odot \vp_{i}$, where $\vp_{i}$ denotes a backdoor trigger pattern, which can either be specific to each sample or consistent across samples (in this case, $\vp_{1}=\vp_{2}=\cdots=\vp_{|\gD_{p}|}$). The operator $\odot$ denotes an element-wise modification process, such as addition, subtraction, or replacement. Typically, all compromised samples within $\gD_{p}$ share a common backdoor target label $y'$.

Training a model on the (potentially) manipulated dataset $\gD'$ can be considered as a dual-task learning problem involving: 1) a ``clean task" focusing on the clean subset of data $\gD_c$, and 2) a ``backdoor task" concentrating on the poisoned subset of data $\gD_p$. In standard (unsecured) training, the model is trained on both the clean and poisoned data by minimizing the following empirical error:
\begin{equation}
\label{eq:backdoor}
     \gL = \underbrace{\sE_{(\vx, y) \sim \gD_c} [\ell(f_{\theta}(\vx), y)]}_{\textup{clean task}} + \underbrace{\sE_{(\vx', y') \sim \gD_p} [\ell(f_{\theta}(\vx'), y')]}_{\textup{backdoor task}}.
\end{equation}
The outcome of the above optimization is a backdoored model, denoted as $f_{\theta}'$, which consistently outputs the backdoor target label whenever the trigger pattern appears, i.e., $f_{\theta}'(\vx \odot \vp) = y'$.

To inhibit the learning of backdoor samples, ABL employs its first stage to segregate clean samples into \Scale[0.85]{\hat{\gD}_c} and possibly poisoned samples into \Scale[0.85]{\hat{\gD}_p}. It then seeks to minimize the model's error on \Scale[0.85]{\hat{\gD}_c} while maximizing its error on \Scale[0.85]{\hat{\gD}_p} in its second stage by minimizing the following objective:
\begin{equation}\label{eq:abl}
\gL = \underbrace{\sE_{(\vx, y) \sim {\hat\gD}_c} [\ell(f_{\theta}(\vx), y)]}_{\textup{clean training}} - \underbrace{\sE_{(\vx', y') \sim {\hat\gD}_p} [\ell(f_{\theta}(\vx'), y')]}_{\textup{backdoor unlearning}}.
\end{equation}
Notably, the opposite optimization direction on \Scale[0.85]{\hat{\gD}_p} can help unlearn the backdoor trigger from the model.
Following ABL, we next introduce our E2ABL method, which leverages an enhanced optimization objective with the integration of a second classification head.

\subsection{Proposed E2ABL Method}

\begin{figure}[h!]
\includegraphics[width=\columnwidth]{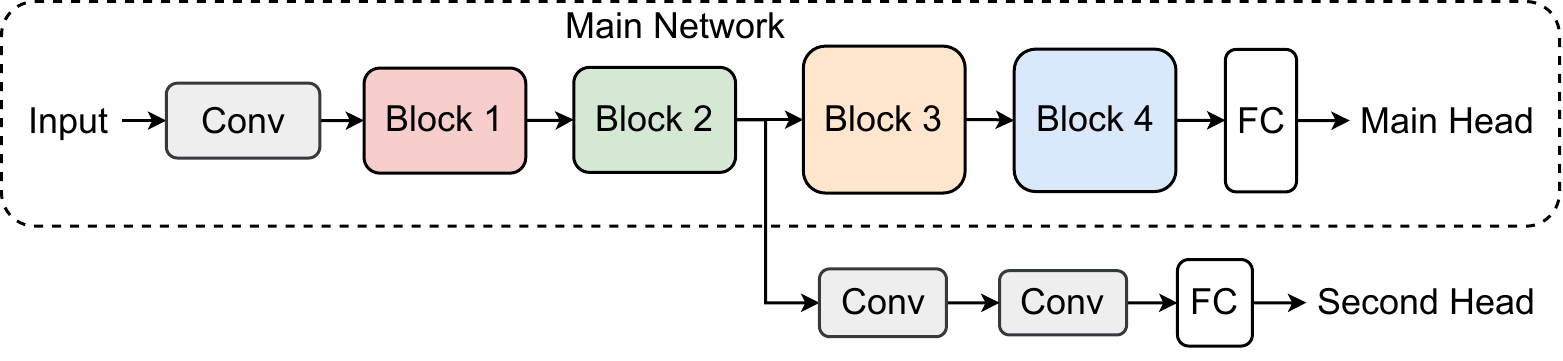}
\centering
\vspace*{-3ex}
\caption{The second classification head attached to ResNet-34.}
\vspace*{-1ex}
\label{fig:second-head}
\end{figure}

\paragraph{Motivation} As discovered in ABL \cite{li2021anti}, a strong correlation exists between the trigger pattern and the backdoor label. The more potent the attack, the stronger the correlation becomes, which in turn allows the backdoor samples to be learned more quickly by the model. This indicates that the backdoor task, as defined in \Eqref{eq:backdoor}, is generally less complex than the clean task.
This suggests that a simpler task could be efficiently learned by either a smaller network or the shallow layers of a deeper network. Such insight motivates us to add a second classification head to the shallow layers of the target model, specifically designed to learn and trap the backdoor samples. The placement of the second head in a ResNet-34 model \cite{he2016deep} is illustrated in Figure \ref{fig:second-head}.

Leveraging the second head, our E2ABL methodology trains a model by minimizing the following empirical error:
\begin{align}\label{eq:e2abl}
    \gL &= \underbrace{\sE_{(\vx, y) \sim \hat{\gD}_c} [\ell(h_1(\vx), y)]}_{\textup{main head: clean leaning}} + \underbrace{\sE_{(\vx', y') \sim \hat{\gD}_p} [\ell(h_2(\vx'), y')]}_{\textup{second head: backdoor learning}} \nonumber \\
    & + \underbrace{\sE_{(\vx', y_\textup{rectified}) \sim {\gD}^*} [\ell(h_1(\vx'), y_\textup{rectified})]}_{\textup{main head: backdoor recovery}},
\end{align}
where \Scale[0.85]{h_1(\cdot)} and \Scale[0.85]{h_2(\cdot)} denote the primary and second heads of the network $f_{\theta}$ respectively. The detected clean and poisoned samples are denoted as \Scale[0.85]{\hat{\gD}_c} and \Scale[0.85]{\hat{\gD}_p} (maintaining \Scale[0.85]{\hat{\gD}_c \cap \hat{\gD}_p =\emptyset} and \Scale[0.85]{\gD' = \hat{\gD}_c \cup \hat{\gD}_p}). And, we use \Scale[0.85]{\gD^*} to denote samples from the detected poison set \Scale[0.85]{\hat{\gD}_p} with \textbf{corrected} class labels.

In our design, the second head plays a crucial role in enhancing the robustness of the training process. But it will be removed when training is completed, i.e., only the network and the main head will be used as the final model ($f=h_1$). A noteworthy distinction from the ABL model, as outlined in \Eqref{eq:abl}, is that the second head's objective is to minimize the error on \Scale[0.85]{\hat{\gD}_p} (as opposed to maximizing it). This is to effectively detect and trap backdoor samples to the second head.
Moreover, E2ABL utlizes a specialized detection and recovery strategy for both \Scale[0.85]{\hat{\gD}_c} and \Scale[0.85]{\hat{\gD}_p}. This involves a method premised on the drop rate of training loss for detecting backdoor samples and the subsequent recovery of their true classes. The specifics of this approach will be described in subsequent sections.

\subsubsection{Backdoor Sample Detection}

Through empirical observation, we note that several convolutional channels within the shallow layers are closely tied to the backdoor trigger. These channels consistently produce specific features for nearly every backdoored input, which is so even for dynamic backdoor attacks that utilize sample-wise trigger patterns.
The deep layers of the model will amplify these features, prompting the model to predict the backdoor class. More significantly, due to their short-cut nature \cite{geirhos2020shortcut}, these trigger-related features can be rapidly and adequately learned even with very shallow networks \cite{li2022deep,yang2022backdoor}. In light of this, E2ABL utilizes an additional classification head to capture the backdoor features at the shallow layers and trap those features to safeguard the main head during the training procedure.

Specifically, the second head is designed to learn backdoor features at several shallow layers, constituting two new convolutional layers and one fully connected (FC) layer, as depicted in Figure \ref{fig:second-head}. The final output of this second head aligns with that of the main head, producing class probabilities.
The pathway from the input to the output of the second head operates as a shallow model, adept at learning the backdoor features. Prior to training the main network or executing backdoor sample detection, the second head first undergoes a self-training phase (lasting for only a few epochs) on the entire dataset \Scale[0.85]{\gD'}, as a warm-up.
Subsequently, it partitions all samples into two subsets: \Scale[0.85]{\hat{\gD}_c} and \Scale[0.85]{\hat{\gD}_p}, based on the loss reduction during the warm-up phase. Samples that exhibit the most precipitous decline in loss are allocated to the poison subset, \Scale[0.85]{\hat{\gD}_p}. The metric used to bifurcate the training data is defined as follows:
\begin{equation}
\label{eq:sum_loss_drop}
     \Delta\ell = \sum_{i=2}^n \ \frac{\mathcal{L}_{i} - \mathcal{L}_{i-1}}{(i-1)^2},
\end{equation}
where $\mathcal{L}_{i}$ is the historical loss value in the $i^{th}$ epoch. Following the poisoning rate (less than 20\%) assumption made in ABL, we segregate the top 20\% of training samples (more discussions on this percentage are given in the ablation studies section), exhibiting the most significant loss drops $\Delta\ell$, into \Scale[0.85]{\hat{\gD}_p}, while the remainder is retained in \Scale[0.85]{\hat{\gD}_c}. These two subsets form the basis for training the main and second heads as detailed in \Eqref{eq:e2abl}. This detection process is performed at the conclusion of each training epoch, subsequent to the initial warm-up phase.

Our backdoor sample detection strategy as described above is notably simpler than the loss-restricted training and filtering strategy employed in ABL. Unlike ABL which strives to accurately identify the backdoor samples at this stage, our method partitions the dataset into two broad subsets. While it is likely that the detected poison subset \Scale[0.85]{\hat{\gD}_p} will encompass the majority of the backdoor samples (as some clean ``easy-to-learn" samples could also have exceptionally high loss drop, as demonstrated in \cite{li2021anti}), we cannot fully separate the backdoor samples from the rest at this stage. Following this, we will introduce the second key technique of E2ABL that makes it work more effectively than ABL.


\subsubsection{True Class Recovery}\label{sec:true_label_recovery}
This operation purifies the labels of certain samples in the detected poison subset \Scale[0.85]{\hat{\gD}_p} and incorporates these corrected samples into an additional subset, \Scale[0.85]{\gD^{*}}. This additional subset is then used to train the main head in conjunction with \Scale[0.85]{\hat{\gD}_c}. 
Backdoor attacks are commonly tied to a specific label deliberately chosen by the adversary. This particular label is referred to as the backdoor label (or class). Intuitively, if we manage to successfully identify and modify this backdoor label, we could effectively break the correlation between the trigger pattern and the backdoor label, thereby mitigating the attack's impact. Furthermore, through empirical experiments, we discovered that the true label of a backdoor sample can be recovered using the output of the main head. As the main head remains unaffected by the attack during training, it allows us to recover the authentic identity of the sample, ensuring the model learns the correct information.

Intuitively, samples in \Scale[0.85]{\hat{\gD}_p} with the largest loss drop, such as the top 1\% of all training samples, are most likely to be backdoor samples. We take their predominant label prediction (the one with the highest softmax value) from the second head as the backdoor label. This detection strategy is built upon two assumptions: 1) backdoor samples have the most significant loss drop, and 2) the second head is designed to be highly skilled in learning the backdoor. Following this phase, samples with the most notable loss drop are relabeled based on the main head's prediction and incorporated into the \Scale[0.85]{\gD^{*}}.


The entire training procedure of E2ABL is described in Algorithm \ref{alg:training}. Throughout the training process, the second head functions as a backdoor supervisor, segregating the input samples into clean and backdoor subsets, identifying the most suspicious samples within the backdoor subset and correcting their labels, and discovering samples from the backdoor subset that are not particularly suspicious (those bearing a non-backdoor label). These operations are all contingent on the predictions made by the second head. In summary, both heads are concurrently trained on the corresponding subsets defined in \Eqref{eq:e2abl}. These subsets are dynamically updated at the end of each epoch, after the warm-up of the second head.




\begin{algorithm}[]
\caption{Training Procedure of E2ABL}
\label{alg:training}
\vspace{0.1ex}

\begin{flushleft}

{\bfseries Input:} 
\Scale[0.85]{\gD'}: backdoor-poisoned dataset; \Scale[0.85]{h_1(\cdot)}, \Scale[0.85]{h_2(\cdot)}: the main and second head; \Scale[0.85]{\ell_\text{CE}(\cdot)}: cross-entropy loss; \Scale[0.85]{\hat{\gD}_c}: detected clean subset; \Scale[0.85]{\hat{\gD}_p}: detected poison subset; \Scale[0.85]{\gD^*}: samples with corrected labels. \\

{\bfseries Output:}
$h_1(\cdot)$\\

{\bfseries Hyper-parameters:} \Scale[0.85]{T_\text{warm\_start}}: warm start epochs for the second head; $T_\text{training}$: total training epochs; $\gamma$: clean percentage of samples for detection; $y_{\textup{rectified}}$: the non-backdoor class with the maximum probability. 

\end{flushleft}

\begin{algorithmic}[1]
\FOR{$i$ in [1, \ldots,\Scale[0.85]{T_\text{warm\_start}}]}
\STATE \# Warm up the second head
\STATE $h_2 \leftarrow \argmin_{h_2} \E_{(\vx,y) \sim \gD'} [\ell_\text{CE}(h_2(\vx), y)]$
\ENDFOR
\FOR{$i$ in [1, \ldots, $T_\text{training}$]}

\STATE \# Detect backdoor samples and fix their labels

\STATE \Scale[0.85]{\hat{\gD}_c} $\leftarrow $ for $\gamma_c\%$ low loss drop samples in \Scale[0.85]{\gD'} w.r.t. \Scale[0.85]{h_2(\cdot)} \\

\STATE \Scale[0.85]{\hat{\gD}_p} $\leftarrow $ for $\gamma_p\%$ high loss drop samples in \Scale[0.85]{\gD'} w.r.t. \Scale[0.85]{h_2(\cdot)} \\

\STATE \Scale[0.85]{\gD^{*}} $\leftarrow ( \Scale[0.85]{\gD' \setminus \hat{\gD}_c}) $ \ with corrected labels $y_{\textup{rectified}}$

\STATE \# Update the two heads
\STATE $h_1 \leftarrow \argmin_{h_1} \E_{(\vx,y) \sim \hat{\gD}_c} [\ell_\text{CE}(h_1(\vx), y)]$
\STATE $h_1 \leftarrow \argmin_{h_1} \E_{(\vx',y_{\textup{rectified}}) \sim \gD^{*}} [\ell_\text{CE}(h_1(\vx'), y_{\textup{rectified}})]$
\STATE $h_2 \leftarrow \argmin_{h_2} \E_{(\vx',y') \sim \hat{\gD}_p} [\ell_\text{CE}(h_2(\vx'), y']$
\ENDFOR
\vspace{0.2ex}
\STATE \textbf{return} $h_1$
\end{algorithmic}
\end{algorithm}

\section{Experiments}
\label{sec:experiment}

\begin{table*}[t!]
\centering
\setlength{\tabcolsep}{0.8em}
\renewcommand{\arraystretch}{1.34}

\caption{The attack success rate (ASR) (lower is better) and the clean accuracy (CA) 
 (higher is better) of 4 backdoor defense methods against state-of-the-art backdoor attacks on both image and time series datasets. `None' means no attack.}
\scalebox{1.06}{
\begin{tabular}{c|c|cc||cc|cc|cc|cc}
\toprule
\multirow{2}{*}{Dataset} & \multirow{2}{*}{Attack} & \multicolumn{2}{c||}{No Defense} & \multicolumn{2}{c|}{FP} & \multicolumn{2}{c|}{NAD} & \multicolumn{2}{c|}{ABL} & \multicolumn{2}{c}{\textit{E2ABL (Ours)}} \\
 &  & ASR & CA & ASR & CA & ASR & CA & ASR & CA & ASR & CA \\ \hline \hline
\multirow{11}{*}{CIFAR-10} & None & N/A & 89.32\% & N/A & 86.07\% & N/A & 87.43\% & N/A & 88.04\% & N/A & \textbf{89.39\%} \\ \cline{2-12}
 & BadNets & 100.0\% & 87.51\% & 99.87\% & 82.90\% & 3.48\% & 84.11\% & 3.18\% & 86.44\% & \textbf{0.17\%} & \textbf{87.96\%}\\
 & Blend & 100.0\% & 85.64\% & 86.40\% & 82.16\% & \textbf{4.97\%} & 83.11\% & 16.85\% & 84.93\% & 8.95\% & \textbf{85.21\%} \\
 & Trojan & 100.0\% & 88.77\% & 65.17\% & 82.46\% & 16.43\% & 76.59\% & 3.45\% & 87.38\% & \textbf{1.87\%} & \textbf{88.26\%} \\
 & Dynamic & 100.0\% & 86.40\% & 87.63\% & 82.48\% & 31.59\% & 73.14\% & 18.83\% & 85.93\% & \textbf{12.15\%} & \textbf{86.22\%} \\ \cline{2-12}
 & CL & 99.81\% & 84.11\% & 51.94\% & 82.16\% & 14.95\% & 81.14\% & \textbf{0.00\%} & 89.05\% & 0.18\% & \textbf{89.11\%} \\
 & SIG & 99.45\% & 84.58\% & 74.81\% & 83.04\% & 2.37\% & 82.18\% & \textbf{0.08\%} & 88.44\% & 0.25\% & \textbf{88.92\%} \\ \cline{2-12}
 & LBA & 99.02\% & 82.89\% & 56.72\% & 81.19\% & 10.07\% & 78.28\% & 0.12\% & 81.26\% & \textbf{0.03\%} & \textbf{82.34\%} \\
 & CBA & 89.14\% & 85.71\% & 75.94\% & 81.32\% & 34.94\% & 81.12\% & 29.28\% & 84.75\% & \textbf{25.64\%} & \textbf{85.27\%} \\
 & DFST & 99.55\% & 84.92\% & 78.47\% & 81.67\% & 35.01\% & 79.39\% & 5.47\% & 81.14\% & \textbf{3.21\%} & \textbf{81.95\%} \\ \cline{2-12}
 & \textbf{Average} & 98.55\% & 85.61\% & 75.22\% & 82.15\% & 17.09\% & 79.90\% & 8.58\% & 85.48\% & \textbf{5.83\%} & \textbf{86.14\%} \\\hline \hline
 \multirow{7}{*}{GTSRB} & None & N/A & 97.91\% & N/A & 90.48\% & N/A & 95.64\% & N/A & 96.78\% & N/A & \textbf{97.95\%} \\ \cline{2-12}
 & BadNets & 100.0\% & 97.50\% & 99.40\% & 88.12\% & 0.22\% & 89.62\% & \textbf{0.05\%} & 96.42\% & 0.09\% & \textbf{96.89\%} \\
 & Blend & 100.0\% & 96.12\% & 99.18\% & 87.34\% & \textbf{7.54\%} & 93.16\% & 25.81\% & 93.27\% & 12.18\% & \textbf{93.95\%} \\
 & Trojan & 99.84\% & 96.74\% & 93.41\% & 85.72\% & 0.46\% & 90.55\% & 0.43\% & 95.24\% & \textbf{0.27\%} & \textbf{95.68\%} \\
 & Dynamic & 100.0\% & 97.13\% & 99.82\% & 85.16\% & 69.64\% & 79.15\% & 6.48\% & 95.87\% & \textbf{5.69\%} & \textbf{96.23\%} \\ \cline{2-12}
 & SIG & 96.58\% & 97.02\% & 81.04\% & 86.43\% & 4.97\% & 90.42\% & 5.45\% & 96.41\% & \textbf{4.78\%} & \textbf{96.79\%} \\ \cline{2-12}
 & \textbf{Average} & 99.28\% & 96.90\% & 94.57\% & 86.55\% & 16.57\% & 88.58\% & 7.64\% & 95.44\% & \textbf{4.60\%} & \textbf{95.91\%} \\ \hline \hline
 \multirow{5}{*}{ArabicDigits} & None & N/A & 86.24\% & N/A & 82.15\% & N/A & 83.44\% & N/A & 84.95\% & N/A & \textbf{86.10\%} \\ \cline{2-12}
 & TT-FGSM & 83.43\% & 72.27\% & 21.14\% & 62.78\% & 7.63\% & 69.48\% & \textbf{0.04\%} & 83.56\% & 0.13\% & \textbf{84.32\%} \\
 & TT-DE & 96.06\% & 69.12\% & 42.83\% & 60.46\% & 26.82\% & 68.17\% & 4.16\% & 81.83\% & \textbf{2.54\%} & \textbf{82.85\%} \\
 & TSBA-B & 97.70\% & 83.49\% & 63.22\% & 62.15\% & 57.63\% & 71.27\% & 24.11\% & 79.19\% & \textbf{16.89\%} & \textbf{81.52\%} \\ \cline{2-12}
 & \textbf{Average} & 92.40\% & 74.96\% & 42.40\% & 61.80\% & 30.69\% & 69.64\% & 9.44\% & 81.53\% & \textbf{6.52\%} & \textbf{82.90\%} \\ \hline \hline
\multirow{5}{*}{ECG5000} & None & N/A & 99.60\% & N/A & 95.50\% & N/A & 96.90\% & N/A & 97.40\% & N/A & \textbf{99.40\%} \\ \cline{2-12}
 & TT-FGSM & 76.10\% & 88.20\% & 20.00\% & 64.00\% & 8.10\% & 72.60\% & \textbf{0.00\%} & 92.90\% & \textbf{0.00\%} & \textbf{96.40\%} \\
 & TT-DE & 98.20\% & 86.40\% & 29.40\% & 63.10\% & 18.20\% & 70.40\% & 1.90\% & 92.60\% & \textbf{0.60\%} & \textbf{96.00\%} \\
 & TSBA-B & 98.70\% & 98.10\% & 58.70\% & 63.50\% & 45.20\% & 74.60\% & 10.80\% & 91.70\% & \textbf{8.60\%} & \textbf{94.80\%} \\ \cline{2-12}
 & \textbf{Average} & 91.00\% & 90.90\% & 36.03\% & 63.53\% & 23.83\% & 72.53\% & 4.23\% & 92.40\% & \textbf{3.07\%} & \textbf{95.73\%} \\ \hline \hline
\multirow{5}{*}{UWave} & None & N/A & 92.47\% & N/A & 86.43\% & N/A & 88.96\% & N/A & 90.17\% & N/A & \textbf{92.54\%} \\ \cline{2-12}
 & TT-FGSM & 87.15\% & 81.10\% & 14.49\% & 72.37\% & 5.62\% & 76.14\% & \textbf{0.37\%} & 88.72\% & 0.69\% & \textbf{91.55\%} \\
 & TT-DE & 96.64\% & 78.12\% & 28.41\% & 71.68\% & 11.27\% & 74.40\% & 3.32\% & 86.62\% & \textbf{1.73\%} & \textbf{91.15\%} \\
 & TSBA-B & 94.13\% & 89.67\% & 54.73\% & 73.69\% & 48.49\% & 77.13\% & 15.34\% & 84.76\% & \textbf{14.64\%} & \textbf{89.27\%} \\ \cline{2-12}
 & \textbf{Average} & 92.64\% & 82.96\% & 32.54\% & 72.58\% & 21.79\% & 75.89\% & 6.34\% & 86.70\% & \textbf{5.69\%} & \textbf{90.66\%} \\
\bottomrule
\end{tabular}
}
\label{tbl:experiment_results}
\end{table*}

\subsection{Attack Configurations}

Our analysis encompasses 9 backdoor attacks on image datasets, including CIFAR-10 \cite{krizhevsky2009learning} and GTSRB \cite{stallkamp2011german} datasets.
The evaluated attacks consist of 4 classic backdoor attacks, including BadNets \cite{gu2017badnets}, Blend \cite{chen2017targeted}, Trojan \cite{liu2018trojaning}, and Dynamic \cite{nguyen2020input}; 2 clean-label backdoor attacks, including Clean-Label attack (CL) \cite{turner2019clean} and Sinusoidal signal attack (SIG) \cite{barni2019new}; and 3 feature-space backdoor attacks, including Latent Backdoor Attack (LBA) \cite{yao2019latent}, Composite Backdoor Attack (CBA) \cite{lin2020composite}, and DFST \cite{cheng2021deep}.
Furthermore, we tested our E2ABL against 3 time series backdoor attacks, including TimeTrojan-FGSM \cite{ding2022towards}, TimeTrojan-DE \cite{ding2022towards}, and TSBA-B \cite{jiang2022backdoor}, using 3 multivariate signal datasets. These datasets include ArabicDigits, ECG5000, and UWave, all sourced from the MTS Archive \cite{UCRArchive}.
All the tested image and time series-based attacks adopt a consistent poison rate of 10\%, with all other training parameters being set as per their original configurations. It is worth noting that several attacks, including CL, LBA, CBA, and DFST, could not be reproduced on certain image datasets such as GTSRB. Consequently, those experiments have been excluded from our reported results.

\subsection{Defense and Training Details}

We assess our E2ABL in comparison with 3 representative defense methods: Fine-pruning (FP) \cite{liu2018fine}, Neural Attention Distillation (NAD) \cite{li2021neural}, and Anti-backdoor Learning (ABL) \cite{li2021anti}. Given that ResNet has been demonstrated to be an effective baseline method for time series classification, as supported by \cite{wang2017time, jiang2022backdoor}, we utilize ResNet-34 as the backbone model for all poisoned datasets in each attack scenario, on both image and time series data.

For FP, we prune the final convolutional layer of each model until the CA falls below the minimum CA under the no-defense condition. In terms of NAD, we follow the standard distillation procedure, which necessitates fine-tuning the backdoored student network for 10 epochs with a 5\% clean data subset. Regarding the original ABL defense, we train the model for 20 epochs, applying a learning rate of 0.1 on CIFAR-10 and 0.01 on GTSRB prior to the turning epoch. We followed the ABL work and set the backdoor isolation and unlearning rate ($\gamma_p$) as 1\%. After successfully segregating 1\% of the potential backdoor samples, we proceed to fine-tune the model for 60 additional epochs with all training samples to restore the model's clean accuracy. We then carry out backdoor unlearning with the 1\% isolated backdoor samples, applying a learning rate of 0.0001 for 20 epochs.

As for our E2ABL defense, we follow the training procedure outlined in Algorithm \ref{alg:training}. We initially trained the second head on the full training dataset for 2 epochs with a learning rate of 0.1 and monitored the loss reduction for each training sample. The subsets defined in \Eqref{eq:e2abl} are dynamically updated based on the weighted sum of the total loss reductions specified in \Eqref{eq:sum_loss_drop}, with the loss drop threshold $\gamma_c$ set as 80 and $\gamma_p$ set as 1. The E2ABL model undergoes a total of 60 epochs of training, inclusive of the 2 warm-up epochs, with a learning rate of 0.01 for the main head and 0.005 for the second head.

\subsection{Effectiveness of E2ABL Defense}

\subsubsection{Comparison to Existing Defenses}

Table \ref{tbl:experiment_results} shows that our E2ABL achieves the best clean accuracy among all backdoor defense techniques while maintaining an exceptionally low Attack Success Rate (ASR) against state-of-the-art backdoor attacks. In the case of clean-label attacks, E2ABL fully recovers clean accuracy while maintaining an almost negligible ASR. On average, our E2ABL outperforms the original ABL by a margin of 2.76\% in terms of lower ASR and 0.66\% in higher Clean Accuracy (CA) across all 9 experiments on the image datasets. For time series, compared to the original ABL, our E2ABL achieves a lower ASR by 1.58\% and a higher CA by 2.89\% on average, spanning all time series experiments.

Previous research has shown that implementing backdoor defense methods on clean training datasets can negatively impact the clean accuracy of the final model \cite{li2021anti}. However, compared to existing defenses, our E2ABL achieved an even higher CA when the training data is completely clean, as shown in the `None' rows for each dataset in Table \ref{tbl:experiment_results}. Interestingly, on certain datasets such as CIFAR-10 and UWave, the models trained by our E2ABL exhibit higher CAs than those trained using standard training procedures. For instance, when the training data is clean, a standard model achieves a CA of 89.32\%, but a model trained with our E2ABL reaches a CA of 89.39\%. This phenomenon may be attributed to the exclusion of those ``easy-to-learn" samples, which could have a negative impact on the model's overall performance. This underlines a unique advantage of our defense method in real-world scenarios where the presence of a backdoor attack remains uncertain.
Note that our defense requires no prior knowledge of the attack, whereas it uses an additional backdoor detection head (i.e., the second head) to determine whether there are any backdoor samples in the training set and recover the potential backdoor label. 

\subsubsection{Effectiveness with Different Subset Sizes}

\begin{figure}[ht]
\includegraphics[width=1.05\columnwidth]{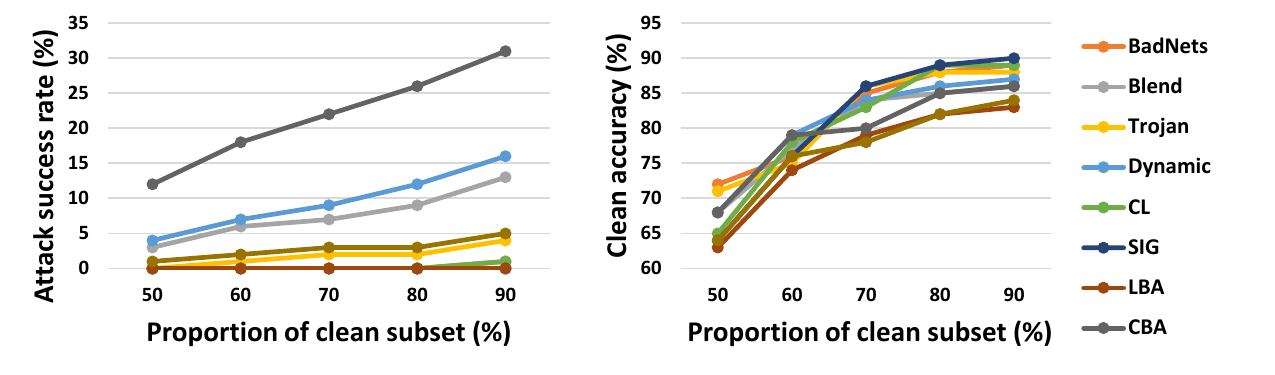}
\centering

\caption{Performance of E2ABL regarding ASR and CA with different isolation rate ($\Scale[0.85]{\gamma_c}$) on CIFAR-10.}

\label{fig:isolation_rate_performance}

\end{figure}

We also study the correlation between the isolation size (the ratio between the isolated clean subset and the full set, $\Scale[0.85]{\gamma_c}$) and the performance of our E2ABL. We test E2ABL with the clean subset size varying from 50\% to 90\% against all the nine state-of-the-art backdoor attacks on the CIFAR-10 dataset and show their ASR and CA results in Figure \ref{fig:isolation_rate_performance}. 
It shows that, with a higher clean subset size, more training samples will be used to train the main head, resulting in moderately higher CA. However, achieving a perfect separation between backdoor and clean samples is not feasible, increasing the clean subset size will reduce the precision of clean sample detection so that poisoned samples have more chance to be mixed into the clean subset, causing a significant rise in ASR (indicates worse performance). We also find that even with a few ($<5$) poisoned samples mixed into the clean subset, a noticeable increase in ASR will be observed in the final model. This also indicates the importance of our proposed strategy that first performs a less accurate but more secure clean-vs-poison isolation and then gradually refines the samples in the poison subset \Scale[0.85]{\hat{\gD}_p} to improve the clean performance.

\subsection{Detection and Recovery Performances}

During the E2ABL procedure, it can distinguish between clean and backdoor samples with high precision. Samples present in \Scale[0.85]{\hat{\gD}_c} were classified as clean samples. However, not all samples in \Scale[0.85]{\hat{\gD}_p} should be considered as backdoor samples. Conceptually, we could designate the 1\% of samples in \Scale[0.85]{\hat{\gD}_p} that have the largest loss drop rate $\Delta\ell$ as the ``most probable" backdoor samples. In this subsection, we delve into the precision of these two subsets to offer further insights into their detection performance.

\begin{table}[h]
\centering
\setlength{\tabcolsep}{0.6em}
\renewcommand{\arraystretch}{1.2}

\caption{Performance of E2ABL in differentiating between clean and backdoor samples, and restoring true labels. The second column shows the precision of identifying backdoor-infected samples within the subset characterized by the top 1\% of loss reductions. The results are computed at the 10$^{th}$ epoch, following the warm start of the second head.}
\scalebox{1.0}{
\begin{tabular}{l|c|c|c|c}
\toprule
 Attack & \begin{tabular}[c]{@{}c@{}}Precision\\ Clean \Scale[1]{\hat{\gD}_c}\end{tabular} & \begin{tabular}[c]{@{}c@{}}Precision\\ Backdoor$_{\text{top}1\%}$\end{tabular} & \begin{tabular}[c]{@{}c@{}}Recall\\ Backdoor \Scale[1]{\hat{\gD}_p}\end{tabular} & \begin{tabular}[c]{@{}c@{}}Precision\\ True Label\end{tabular} \\ \midrule
(1) BadNets & 100\% & 99.8\% & 100\% & 83.8\% \\
(2) Blend & 99.0\% & 98.4\% & 99.6\% & 76.2\% \\
(3) Trojan & 100\% & 99.6\% & 100\% & 72.0\% \\
(4) Dynamic & 98.4\% & 95.2\% & 94.4\% & 88.6\% \\
(5) CL & 100\% & 100\% & 100\% & 0\% \\
(6) SIG & 100\% & 99.6\% & 100\% & 0\% \\
(7) LBA & 100\% & 99.8\% & 100\% & 68.4\% \\
(8) CBA & 97.2\% & 96.4\% & 91.6\% & 76.8\% \\
(9) DFST & 98.8\% & 98.6\% & 99.8\% & 63.0\% \\ \bottomrule
\end{tabular}
}
\vspace{-2ex}
\label{tbl:prec_result}
\end{table}

\subsubsection{Precision of the Detected Clean and Backdoor Samples}
As demonstrated in Table \ref{tbl:prec_result}, our E2ABL exhibits high precision in separating clean and backdoor samples based on the loss drops captured by the second head. However, it is worth noting that stronger backdoor attacks, such as Dynamic and CBA, result in lower detection precision. This means that a small fraction of backdoor samples may not fall within the 1\% of samples in \Scale[0.85]{\hat{\gD}_p} that have the most substantial loss drops. Furthermore, some are not even captured in the \Scale[0.85]{\hat{\gD}_p} set (which contains top 20\% samples based on loss drops), as illustrated by the third column in Table \ref{tbl:prec_result}. This phenomenon explains the higher (worse) ASR observed in some of the experiments and serves as an area where E2ABL could be further refined.
Generally speaking, our method highlights an effective technique for segregating backdoor samples, thereby allowing E2ABL to train clean models on potentially compromised training data.

\subsubsection{Precision of the Recovered True Class Labels}
Our E2ABL method introduces a dynamic recovery of the true class labels of poisoned training samples during the training process to recover certain poisoned samples in \Scale[0.85]{\hat{\gD}_p} to enhance the clean accuracy of the main head. As shown in the last column of Table \ref{tbl:prec_result}, these recovered labels present high precision for dirty-label and feature-based attacks.
Note that as long as the sample is not a backdoor sample, its loss value with respect to the backdoor label at the second head will be notably high, as the second head does not have the capacity to sufficiently learn the clean task. In the case of clean-label attacks (such as CL and SIG), the backdoor-poisoned samples at the second head will point to the adversary-chosen target. Accordingly, the poisoned samples will be \textbf{corrected} to different (although might be incorrect) class labels other than the backdoor target. This disrupts the correlation between the trigger pattern and the backdoor label, making it more challenging for the main head to learn and recognize the backdoor.

\section{Ablation Studies}

\begin{table}[h]
\centering
\setlength{\tabcolsep}{0.5em}
\renewcommand{\arraystretch}{1.2}
\caption{Ablation studies of E2ABL on CIFAR-10. The full names of the attacks are in Table \ref{tbl:prec_result}. The $\Delta$CA and $\Delta$ASR are calculated based on the E2ABL results in Table \ref{tbl:experiment_results}.}
\scalebox{1.05}{
\begin{tabular}{c|cc|cc|cc}
\toprule
 & \multicolumn{2}{c|}{Unlearn Top 1\%} & \multicolumn{2}{c|}{ With No Control} & \multicolumn{2}{c}{Use Two Models} \\
 & $\Delta$ASR & $\Delta$CA & $\Delta$ASR & $\Delta$CA & $\Delta$ASR & $\Delta$CA \\ \midrule
(1) & +1.01\% & -3.54\% & +7.55\% & -0.62\% & +0.93\% & +0.13\% \\
(2) & +1.64\% & -5.41\% & +14.64\% & -4.13\% & +2.16\% & +0.64\% \\
(3) & +0.83\% & -2.55\% & +9.62\% & -1.59\% & +1.14\% & +0.20\% \\
(4) & +0.96\% & -2.94\% & +12.25\% & -1.41\% & +0.85\% & +0.35\% \\
(5) & +0.56\% & -0.94\% & +2.36\% & -2.15\% & +1.02\% & -0.06\% \\
(6) & +1.12\% & -1.15\% & +4.17\% & +0.05\% & +0.81\% & -0.81\% \\
(7) & +0.98\% & -5.42\% & +7.42\% & +0.11\% & +1.23\% & -0.67\% \\
(8) & +0.51\% & -4.73\% & +6.39\% & -1.04\% & +0.79\% & +0.13\% \\
(9) & +1.26\% & -5.10\% & +8.59\% & -2.98\% & +1.34\% & +0.29\% \\ \midrule
avg. & +0.99\% & -3.53\% & +8.11\% & -1.53\% & +1.14\% & +0.02\% \\
\bottomrule
\end{tabular}
}
\vspace{-2ex}
\label{tbl:diff_result}
\end{table}

\subsection{E2ABL Without Label Correction}

To achieve higher CA and lower ASR, our E2ABL attempts to rectify the labels of the detected backdoor samples, specifically targeting the 1\% of samples in \(\hat{\mathcal{D}}_p\) with the largest loss drops. These corrected samples are subsequently included in the subset \(\mathcal{D}^*\), enabling the main head to learn the clean task from them.
To explore alternative approaches to managing the detected backdoor samples, we conducted two experiments without using the corrected samples. First, instead of training the main head with \(\mathcal{D}^*\), we applied the unlearning operation of the top 1\% high loss-drop samples utilizing the original ABL method (using negative cross-entropy loss defined with respect to the backdoor label).
Additionally, we conducted an experiment that entails training the main head without the corrected samples in \(\mathcal{D}^*\).
As presented in the first two columns of Table \ref{tbl:diff_result}, training the main head with the ``rectified" samples in \(\mathcal{D}^*\) leads to improvements in both ASR and CA. In contrast, when no backdoor control (namely backdoor unlearning and true class recovery) is applied in the main head's training, the ASR increases dramatically, and CA declines against the majority of attacks. This indicates the significance of the true class recovery step in our E2ABL method, emphasizing its role in enhancing accuracy and robustness.
In most cases, the proposed backdoor recovery method not only significantly reduces ASR but also boosts CA, resulting from the recovered true labels.

\subsection{A Second Head or a Second Model?}

Our E2ABL methodology incorporates a secondary head that is attached to the shallow layers of the DNN, aimed at detecting and rectifying backdoor samples. This approach is based on the assumption that the backdoor task is substantially more straightforward than the clean task. Such an implementation will naturally lead to the question: ``Can a second model achieve the same result?" The most significant difference between employing a second model as opposed to a second head lies in whether they share the same shallow layer weights with the main network. In essence, without this weight sharing, attaching a second head to the DNN model is equivalent to using two separate DNN models.

To further explore this concept, we conducted an experiment utilizing an alternative \textbf{two-model} setting. In this setting, one smaller model is exclusively trained to differentiate between clean and backdoor subsets, mirroring the function of the second head in E2ABL. Concurrently, a full ResNet-34 model is trained following the same procedure as \( h_1 \), as outlined in Algorithm \ref{alg:training} of the manuscript. The results, presented in the third column of Table \ref{tbl:diff_result}, reveal that the two-model design can only marginally enhance the CA by 0.02\% with an average 1.14\% decline in ASR.

These findings illuminate that utilizing separate weights might compromise the secondary model's proficiency in detecting and isolating backdoor samples. The underlying cause of this limitation is that the two models are not learned synchronously, and thus their learning pace may vary. The shared layer design in E2ABL ensures a coordinated learning process, maximizing both detection efficiency and correction effectiveness. This illustrates the advantages of employing a second head in comparison to a separate two-model approach.

\subsection{Different Isolation and Recovery Rates}
\begin{table}[h]
\centering
\setlength{\tabcolsep}{0.5em}
\renewcommand{\arraystretch}{1.2}
\caption{Performance of E2ABL under different isolation and recovery rates ($\gamma_{rec}$, $\gamma_{iso}$): $\gamma_{iso}$ is the isolation rate, $\gamma_{rec}$ is the recovery rate. The $\Delta$CA and $\Delta$ASR are calculated w.r.t. the result of our default experiment setting with ($\gamma_{rec}$, $\gamma_{iso}$) = (1\%, 80\%) shown in Table \ref{tbl:experiment_results}.}
\scalebox{1.05}{
\begin{tabular}{c|cc|cc|cc}
\toprule
$\gamma$ & \multicolumn{2}{c|}{$(1\%, 70\%)$} & \multicolumn{2}{c|}{ $(2\%, 80\%)$} & \multicolumn{2}{c}{$(5\%, 80\%)$} \\
 & $\Delta$ASR & $\Delta$CA & $\Delta$ASR & $\Delta$CA & $\Delta$ASR & $\Delta$CA \\ \midrule
(1) & +1.05\% & +0.74\% & -0.02\% & -0.21\% & -0.06\% & -0.36\%  \\
(2) & +3.68\% & +0.92\% & -2.94\% & -0.42\% & -3.17\% & -0.96\%  \\
(3) & +1.22\% & +0.47\% & -0.32\% & -0.09\% & -0.79\% & -0.17\%  \\
(4) & +4.11\% & +0.39\% & -3.25\% & -0.47\% & -7.12\% & -0.41\%  \\
(5) & +0.84\% & +0.28\% & -0.02\% & -0.10\% & -0.08\% & -0.05\%  \\
(6) & +0.59\% & +0.37\% & -0.10\% & -0.15\% & -0.17\% & -0.23\%  \\
(7) & +0.63\% & +1.01\% & +0.01\% & +0.08\% & -0.01\% & +0.00\%  \\
(8) & +3.89\% & +0.87\% & -4.58\% & -0.27\% & -8.12\% & -0.95\%  \\
(9) & +1.20\% & +0.59\% & -0.89\% & -0.39\% & -1.27\% & -0.62\%  \\
\midrule
avg. & +1.91\% & +0.63\% & -1.35\% & -0.22\% & -2.31\% & -0.42\%  \\ \bottomrule
\end{tabular}
}
\label{tbl:diff_result_new}
\end{table}

We perform experiments employing three distinct sets of hyperparameters: isolation rate ($\gamma_{p}$) and recovery rate ($\gamma_{c}$). The findings, as detailed in Table \ref{tbl:diff_result_new}, indicate that our proposed method demonstrates robustness in different hyperparameter settings. The following conclusions can be derived:

1) A higher recovery rate (increasing from \(1\%\) to \(5\%\)) can further diminish ASR, while the CA is primarily preserved. This suggests that calibration of the recovery rate will have a limited adverse effect on the system's overall accuracy.

2) Conversely, a higher isolation rate (increasing from \(10\%\) to \(20\%\)) can lead to an improvement in CA, though it causes a marginal increase in ASR by less than \(2\%\). Importantly, the overall ASR still remains minimal, indicating that the method's ability to defend against attacks is maintained, even when the isolation rate is reduced.

These observations underscore the robustness of E2ABL, confirming that adjustments to these particular hyper-parameters have a controlled impact on the system's performance for both image and time series, thereby offering flexibility in tuning according to specific requirements.

\subsection{Model behavior between two heads}

\begin{figure}[ht]
\includegraphics[width=0.48\columnwidth]{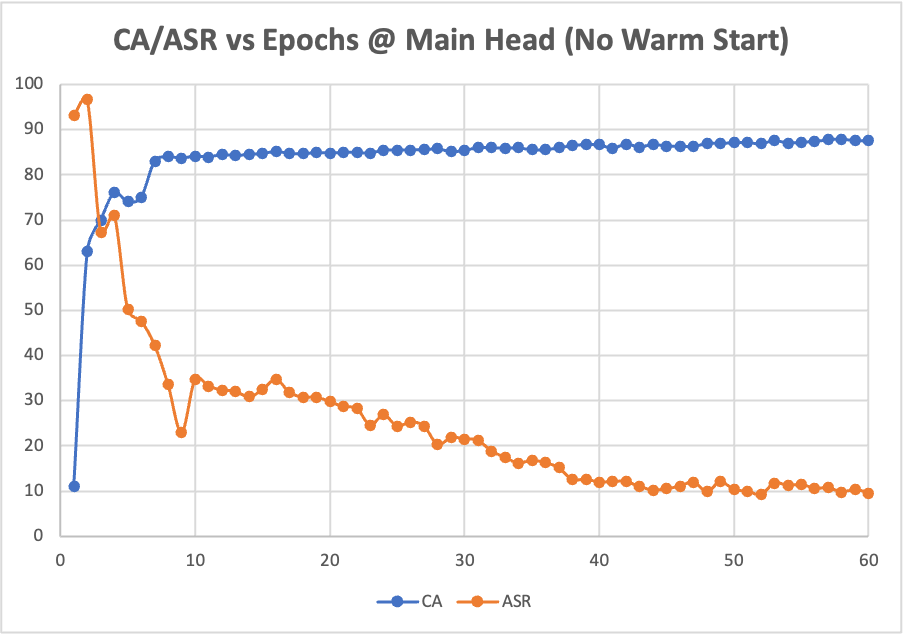}
\includegraphics[width=0.48\columnwidth]{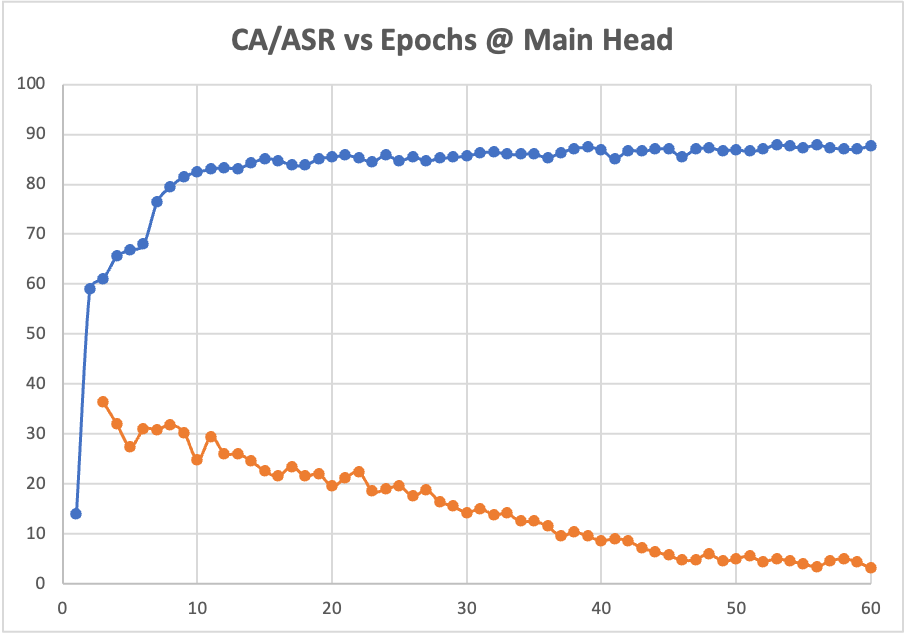}
\includegraphics[width=0.48\columnwidth]{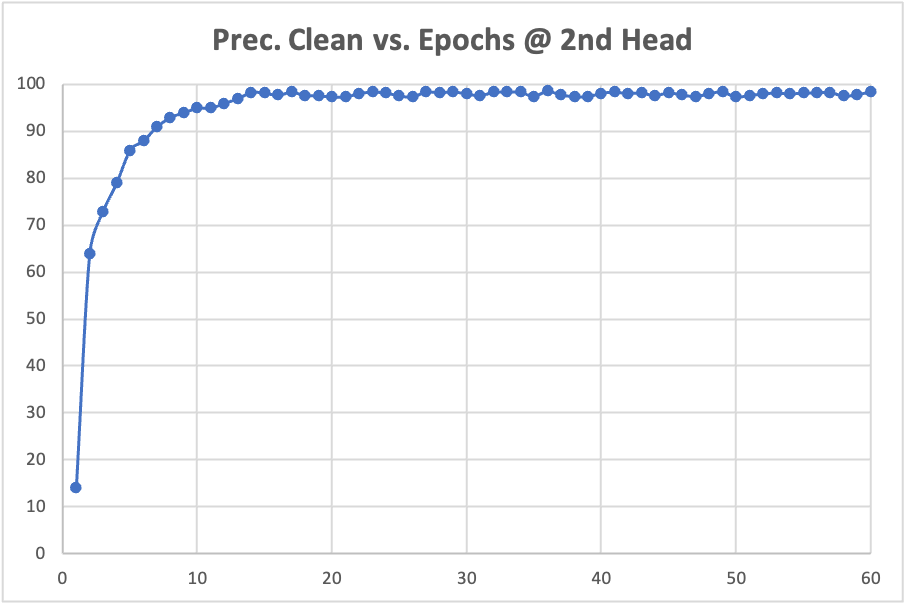}
\includegraphics[width=0.48\columnwidth]{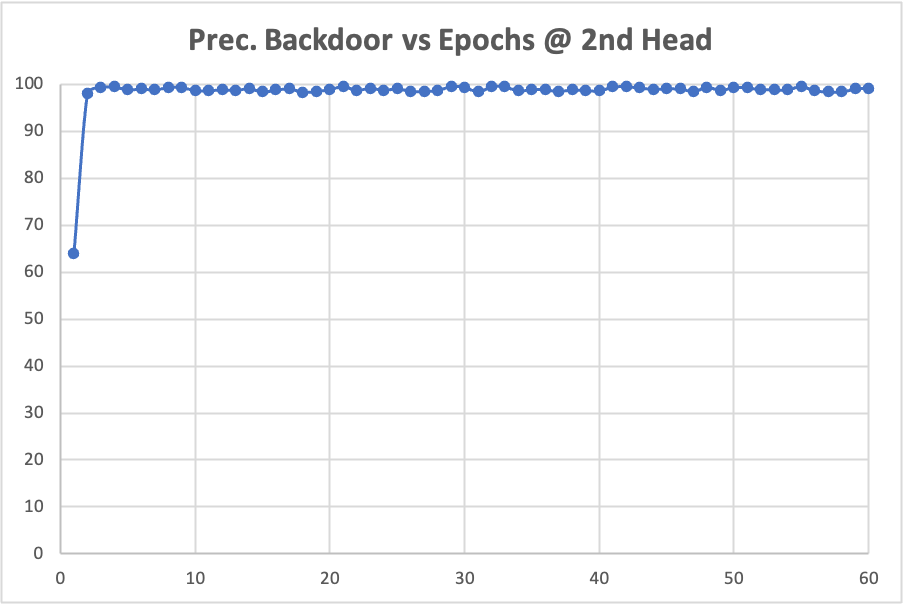}
\centering

\caption{Behavior of dual-head learning over training epochs. The experiments are performed using the CIFAR-10 dataset, incorporating Trojan attacks.}

\label{fig:model_stats_epochs}

\end{figure}

To clearly depict the training behavior of the dual-head model, we carried out supplementary experiments without a warm start (lasting for 2 epochs). The comparative results are presented in the first row of Figure \ref{fig:model_stats_epochs}. We also plotted the precision of clean data and backdoor isolation in the secondary head. The following conclusion can be derived:

1) The cold start approach significantly enhances the second head's ability to effectively segregate backdoor samples. As our E2ABL model is designed to train clean models on poisoned data, only the clean data is channeled into the main head for training purposes, primarily after the removal of the majority of backdoor samples. Simultaneously, these backdoor samples are utilized in the unlearning process.

2) Without the cold start, the final ASR witnesses a 6\% reduction. This outcome is attributed to the exposure of backdoor samples in the early stages of the main head training, which proves challenging to eliminate in the later stages of backdoor unlearning. Conversely, the cold start method first empowers the second head to distinguish between clean and backdoor samples, thereby effectively supporting the subsequent clean training process of the main head.

3) As demonstrated in the bottom row of the figure, the precision of backdoor isolation converges more rapidly compared to clean isolation, reaching nearly 98\% in just 2 epochs. However, the convergence of clean isolation occurs over a longer period, taking nearly 10 epochs. This observation also indicates that the backdoor task is easier to learn compared to the clean task, primarily attributable to the higher loss values incorporated during training.

\subsection{Where To Attach the Second Head?}

In this work, we introduce a secondary classification head, integrated into the shallow layers of the DNN. Functioning as a trap for backdoor samples, this secondary head plays a dual role: it 1) detects these deceptive samples and 2) concurrently corrects their labels. Specifically designed to be sensitive to the presence of backdoors, this innovative secondary head performs essential detection tasks, identifying backdoor samples and striving to recover their true labels. By confining the backdoor samples within the shallow layers, this approach protects the primary head, guiding the model training towards a more secure and trustworthy trajectory.
In our experiment, utilizing ResNet-34 as the backbone model, the secondary head is constructed of two convolutional layers, strategically attached to the termination of the second convolutional stage, as illustrated in Figure \ref{fig:second-head}. It is worth noting that the specific attachment point of the second head and its size (including the number of convolutional layers and the number of neurons in fully connected layers) require further investigation. This assessment can lead to a deeper understanding of the second head, contributing to a more resilient model.

\begin{figure}[h]
\includegraphics[width=1\columnwidth]{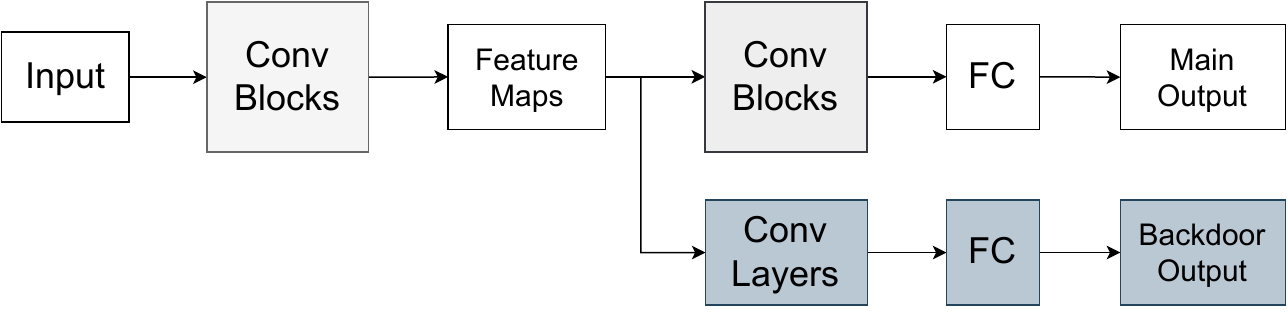}
\centering
\caption{A generalized view of attaching an additional classification head to any DNN models.}
\label{fig:e2abl_model_exp}
\vspace{-1ex}
\end{figure}

As depicted in Figure \ref{fig:e2abl_model_exp}, the entire dual-head model can be understood from the following perspective. First, the early convolutional layers are responsible for extracting high-level feature representations from the input sample. In the case of a backdoor-poisoned sample, these feature representations embody both clean features (which infer the true class of the given input) and backdoor features (which lead the model to misclassify the sample into the target class). Subsequently, these mixed feature representations serve as inputs for both the main head and the secondary head. 
Intriguingly, the two heads are trained with opposing objectives: the main head aims for backdoor-free classification, while the secondary head is sensitized to detect backdoor features. The variations in the attachment point of the secondary head control the depth of the feature representation generated for the latter tasks. 

To investigate the corresponding consequences of varying the attachment point of the secondary head, we conducted controlled experiments with the ResNet-34 model. The location of attachment was systematically altered, ranging from the initial convolutional group (after Block 1) to a position immediately prior to the FC layer (after Block 4). This experimental design allowed us to explore the effects of the secondary head's placement on the model's overall performance and behavior, contributing to our understanding of its optimal integration. 

\begin{table}[h]
\centering
\setlength{\tabcolsep}{0.5em}
\renewcommand{\arraystretch}{1.2}
\caption{Ablation studies of E2ABL on the attachment point of the secondary head. The full names of the attacks can be found in Table 2 of the manuscript. The $\Delta$CA and $\Delta$ASR are calculated based on the CIFAR-10 results in Table 1 of the manuscript (attachment point is after Block 2).}
\scalebox{1.02}{
\begin{tabular}{c|cc|cc|cc}
\toprule
 & \multicolumn{2}{c|}{After Block 1} & \multicolumn{2}{c|}{After Block 3} & \multicolumn{2}{c}{After Block 4} \\
 & $\Delta$ASR & $\Delta$CA & $\Delta$ASR & $\Delta$CA & $\Delta$ASR & $\Delta$CA \\ \midrule
(1) & +1.23\% & +0.61\% & -0.09\% & -1.63\% & +0.98\% & -6.14\% \\
(2) & +1.54\% & +0.69\% & -0.04\% & -2.57\% & +1.12\% & -7.18\% \\
(3) & +2.46\% & -0.92\% & +0.17\% & -3.68\% & -0.14\% & -5.12\% \\
(4) & +3.17\% & +0.80\% & -0.20\% & -6.40\% & +0.68\% & -9.19\% \\
(5) & +0.64\% & -0.56\% & -0.34\% & -6.77\% & +1.63\% & -13.10\% \\
(6) & +0.96\% & +0.75\% & -0.16\% & -4.69\% & +1.32\% & -12.57\% \\
(7) & +1.15\% & -0.64\% & +0.31\% & -5.93\% & +0.65\% & -8.16\% \\
(8) & +0.99\% & -0.61\% & +0.07\% & -7.11\% & +0.84\% & -11.24\% \\
(9) & +1.10\% & +0.73\% & -0.14\% & -5.74\% & +0.79\% & -10.64\% \\ \midrule
avg. & +1.47\% & +0.54\% & -0.05\% & -4.95\% & +0.87\% & -9.26\% \\
\bottomrule
\end{tabular}
}
\vspace{-3ex}
\label{tbl:attach_point}
\end{table}

The results, as presented in Table \ref{tbl:attach_point}, lead to the following conclusion: attaching the secondary head to the output of deeper convolutional groups yields a lower ASR (which is favorable from a defense perspective), but also results in a lower CA (indicating worse performance in prediction). However, when the secondary head is affixed to deeper convolutional groups (such as after Block 4), the dual-head model exhibits a noticeable decline in performance across both metrics, leading to unintended negative consequences. 
The possible cause of this decline in performance may be attributed to the convolutional layers picking up an excessive number of backdoor features, while more benign features are overshadowed or ignored. As a result, the secondary head's capacity to provide protection to the main head in terms of backdoor robustness becomes limited. The current configuration of the dual head model contains $1.2\times$ of parameters with $1.25\times$ run-time compared with the ResNet-34 model.

In our supplementary experiments with alternative backbone models like ResNet-18 and ResNet-50, we noticed consistent trends related to the placement of the secondary head. Generally speaking, our dual-head model tends to achieve an optimal balance between ASR and CA when the secondary head is attached around the midpoint of the convolutional layers.

\section{Conclusion}

In this paper, we proposed the End-to-End Anti-Backdoor Learning (E2ABL) methodology, a simple but innovative technique specifically engineered to train models that remain clean even when exposed to potentially backdoor-poisoned training data. The E2ABL approach, by connecting a second head to the shallow layers of a Deep Neural Network (DNN), serves as a backdoor supervisor that learns, detects, and segregates backdoor samples.
E2ABL also incorporates a partitioning mechanism to distinguish clean samples from potentially poisoned ones, thus creating a subset of backdoor samples. It then employs a novel, dynamic true class recovery process to rectify the labels of a certain proportion of samples within the poisoned subset.
Through extensive experiments on both image and time series data, we have proven E2ABL's effectiveness in defending against 9 backdoor attacks. It can train clean and reliable models even when confronted with sophisticated backdoor attacks. This work presents an actionable solution for safety-critical industries seeking to train models devoid of backdoor vulnerabilities using real-world datasets.
While there are still many open problems, this work has made a first attempt toward a single unified defense for multiple tasks and data modalities. This work can thus serve as a useful baseline for future research.

\bibliographystyle{ieeetr}
\bibliography{ref}

\end{document}